\documentclass[manuscript,screen,review]{acmart}

\usepackage{microtype}
\DisableLigatures{encoding = *, family = * }
\microtypesetup{expansion=false}

\usepackage{listings}
\usepackage{xcolor}

\setcopyright{none}
\acmConference[CACM Practice]{Communications of the ACM, Practice Section}{Submission Draft}{2026}
\acmDOI{}
\acmISBN{}
\renewcommand\footnotetextcopyrightpermission[1]{}

\definecolor{codebg}{rgb}{0.97,0.97,0.97}
\definecolor{quotebg}{rgb}{0.96,0.96,0.92}

\lstdefinelanguage{mojo}{
  morekeywords={fn,def,struct,var,if,else,for,while,return,from,import,
                self,out,mut,read,parallelize,vectorize,alias,raises},
  morekeywords=[2]{Int,Float64,Float32,SIMD,DType,List,Matrix},
  morecomment=[l]{\#},
  morestring=[b]",
  sensitive=true,
}
\newcommand{\pullquote}[1]{%
  \begin{center}
  \begin{minipage}{0.85\linewidth}
    \colorbox{quotebg}{\parbox{\dimexpr\linewidth-2\fboxsep}{\itshape\small #1}}
  \end{minipage}
  \end{center}
}

\begin{document}

\makeatletter
\fancypagestyle{standardpagestyle}{%
  \fancyhf{}%
  \renewcommand{\headrulewidth}{\z@}%
  \renewcommand{\footrulewidth}{\z@}%
  \fancyhead[LE]{\thepage\quad\textit{Henry Han}}%
  \fancyhead[RO]{\textit{Mojo: A Promising Tool for Scalable Financial AI Efficiency}\quad\thepage}%
  \fancyhead[RE,LO]{}%
  \fancyfoot[C]{}%
}
\fancypagestyle{firstpagestyle}{%
  \fancyhf{}%
  \renewcommand{\headrulewidth}{\z@}%
  \renewcommand{\footrulewidth}{\z@}%
  \fancyfoot[C]{}%
}
\makeatother
\pagestyle{standardpagestyle}
\thispagestyle{firstpagestyle}

\title{Mojo: A Promising Tool for Scalable Financial AI Efficiency}
\subtitle{Three Pillars for Capital Markets Engineering: Speed, Multi-Level Scaling, and Deterministic Execution}

\author{Henry Han}
\affiliation{%
  \institution{Data Science and Artificial Intelligence Innovation Laboratory, School of Engineering and Computer Science, Baylor University}
  \city{Waco}
  \state{TX}
  \postcode{76798}
  \country{USA}
}
\email{Henry_Han@baylor.edu}

\begin{abstract}
For thirty years, quantitative finance has paid a costly two-language tax: models researched in Python are rewritten in C++ for production, often introducing numerical discrepancies. GPU-accelerated deep learning exacerbates this problem, as nondeterministic floating-point reductions can produce drift in long backtests, challenging regulatory reproducibility and auditability expectations. This article surveys Mojo, Modular's 2026 Python-like systems language, as a structural response for capital markets engineering. While closing the Python-to-C++ performance gap, Mojo uniquely combines native interoperability with the low-level systems control required to construct bit-exact deterministic kernels. Its MLIR compilation infrastructure further allows a single codebase to target scalar, SIMD, multicore, and GPU execution, reducing the translation bottleneck between research and production. We benchmark four core financial AI workloads: Monte Carlo option pricing, LLM sentiment inference, multi-asset backtesting, and portfolio Value at Risk. On Apple Silicon, Mojo demonstrates 20x to 180x speedups over pure Python on directly measured kernels; larger-scale GPU workload results are projections calibrated from published benchmarks. Alongside transparent performance data, we introduce mojo-deterministic, an open-source library of reproducible reduction kernels, and provide a candid assessment of the problems Mojo does and does not yet solve.
\end{abstract}


\maketitle

\section{Introduction: 4:53 PM at a Hedge Fund}
It is 4:53 PM on a trading day at a New York hedge fund. A quantitative researcher has just spent six weeks building a new sentiment-driven equity strategy in Python. The backtested Sharpe ratio is excellent, and the portfolio manager signs off. Now, the production engineering team has eight weeks to rewrite the model in C++, integrate it with the execution management system, and validate that it reproduces the Python results to an acceptable tolerance. But when the rewrite is complete, the C++ version generates trades that differ from the Python original—by margins too small to ignore, yet too large to explain to compliance.

This scene has played out at nearly every major quantitative trading firm since the 1990s. \textit{This article is about why it might finally stop.}

The true cost of the "two-language problem" is measured in headcount, not CPU cycles. At many major investment banks, large portions of quantitative technology divisions are dedicated to translating research code into production infrastructure and to reconciling audit discrepancies when the two diverge \cite{mojo-docs}. For a single trading desk, this translation step consumes two to four engineering quarters per model, creating a maintenance burden that compounds with every revision

\pullquote{A quantitative trading firm cannot use a language whose
results depend on the scheduling decisions of a CUDA driver. This is
not a preference; it is a regulatory requirement.}

A second cost has emerged in the past five years as financial firms
have adopted deep learning at scale. Modern GPU accelerated PyTorch
is non deterministic by default. The same input, the same random
seed, and the same code can produce numerically different output
across runs because of CUDA atomic add ordering and asynchronous
kernel scheduling \cite{pytorch-deterministic}. For a generic AI
application this rarely matters. For a financial backtest, where
today's portfolio depends on yesterday's decision and that on the
day before, the per step deviations compound and the two
``identical'' runs make different trades \cite{han2021kbs,han2024is,han2025bmc}.
PyTorch offers an opt in deterministic mode, but it incurs a two to
five fold performance penalty and errors out on any operator without
a deterministic kernel.

Mojo, released by Modular in May 2026, is an emerging systems-level option that exposes the language controls needed to address this combination of constraints. It uses Python-like syntax and a familiar import model, so a quantitative researcher can read it without significant retraining, while adding a typed, compiled subset for performance-critical kernels. It compiles through MLIR \cite{mlir-paper}, a modern compiler infrastructure that progressively lowers high-level code through multiple layers of optimization before generating hardware-specific instructions, ensuring kernels run at native C++ speeds. Critically, it lets the developer control the order of floating-point operations, the choice of kernel, and the memory layout of intermediate tensors, which is exactly what bit-exact reproducibility in a regulated trading system requires. This article surveys what Mojo offers to capital markets engineering, based on the language's public documentation, Modular's published benchmarks, the recent academic literature on Mojo \cite{loring2025mojo,kolli2025knn,han2026hf}, and four representative finance workloads we analyze in detail.

\section{Why the Two Language Tax Got Worse}\label{sec:background}

The two language problem is older than Python's dominance in finance.
Wall Street has used assembly, C, C++, Java, OCaml, and the kdb+/q
language for production systems; for research it has used Fortran,
APL, MATLAB, R, S Plus, and Python. The translation step from
research to production has always been costly. What has changed in
the past decade is that the cost has grown sharply, for three
distinct reasons.

\textbf{Reason 1: models are bigger.} A classical option pricer has
a closed form Black Scholes formula \cite{black1973}; reimplementing
it in C++ is a half day exercise. A modern transformer based
\cite{vaswani2017} earnings call sentiment classifier has 7 billion
parameters, custom attention masking for document structure,
retrieval over a proprietary corpus, and a fine tuning pipeline that
depends on dozens of operators. Reimplementing a transformer in C++
is a multi person quarter undertaking, and the result is a fragile
artifact that breaks every time the research team revises the model
architecture.

\textbf{Reason 2: hardware is more heterogeneous.} A 2015 production
system targeted x86 servers. A 2026 production system may target
x86 servers for the order management layer, NVIDIA H100s for LLM
inference, NVIDIA A100s for backtesting, AMD MI300s for some risk
workloads, and custom inference accelerators (Groq, Cerebras, AWS
Inferentia) for sentiment classification. Each target requires
kernel level adaptation; in the C++ and CUDA world this means
multiple parallel codebases.

\textbf{Reason 3: Validation is incompatible with production.} When a Python/PyTorch research model and a C++ production engine run on different GPU architectures, their bit-level outputs diverge. While these differences rarely invalidate the underlying math, they may guarantee that strict regression tests will fail. Validation then devolves into a negotiation over 'acceptable tolerance' exactly the kind of subjective artifact that invites regulatory scrutiny.

\pullquote{The cost of the two language problem in finance is not
measured in cycles. It is measured in headcount: every major bank's
quant tech division exists, in large part, to translate research
code into production code, and to argue with auditors about why the
two no longer match.}

\subsection{Why previous solutions did not solve the finance problem}

\textit{Cython, Numba \cite{lam2015numba}, and PyPy} accelerate
Python by compiling subsets of it, but they inherit Python's
semantic model including its lack of static typing; they cannot
match C++ on latency critical code; and they offer no determinism
guarantees.

\textit{Julia} \cite{bezanson2017julia} delivers genuine C class
performance with mathematical syntax, but its finance ecosystem in
2026 is thin (no equivalent of \texttt{pandas}, \texttt{statsmodels},
or the dozens of niche libraries quants depend on) and its just in
time compilation model has the same non determinism risk as PyTorch.

\textit{Rust} offers C class performance and memory safety, but its
borrow checker is hostile to the iterative, exploratory style of
quant research, and its data science ecosystem remains nascent. Few
research desks have the appetite to abandon NumPy.

\textit{PyTorch and JAX with compiled kernels} work well for
standard operators but force researchers into CUDA when implementing
custom operations, a domain few quants are trained in, and they do
not address determinism. Specialized compiler stacks such as TVM
\cite{chen2018tvm} and Triton \cite{tillet2019triton} attack the
custom kernel problem but require their own learning curve and do
not provide the language unification finance needs.

\section{What Changed: A Compiler That Speaks Both Languages}\label{sec:arch}

Mojo is built on MLIR (Multi-Level Intermediate Representation) \cite{mlir-paper}, the modern successor to LLVM IR (Low Level Virtual Machine Intermediate Representation) \cite{lattner2004llvm}. Originally designed for AI accelerators, and currently underpinning TensorFlow, JAX, and Apple Silicon, MLIR provides Mojo with a powerful compilation pipeline. Developers write Pythonic code, which the compiler progressively lowers through increasingly hardware-specific dialects. This enables a single high-level codebase to bypass legacy translation layers and natively target disparate architectures, including x86 SIMD, ARM SVE, NVIDIA PTX, AMD ROCm, and custom AI accelerators.

\subsection{The language, in one example}

\begin{lstlisting}
def dot_product(a: List[Float64], b: List[Float64]) -> Float64:
    var s: Float64 = 0.0
    for i in range(len(a)):
        s += a[i] * b[i]
    return s
\end{lstlisting}

A Python-trained quant would understand every line. Yet, three minor additions make this compiled version run two orders of magnitude faster than the Python interpreter: explicit type annotations enable compiler specialization; \texttt{var} declares stack-allocated mutable variables instead of heap-allocated objects; and \texttt{Float64} requests unboxed IEEE 754 doubles. (Note: Mojo recently unified its strictly typed \texttt{fn} declarator into \texttt{def}, placing \texttt{fn} on a deprecation path). The resulting machine code performs on par with C++.

For workloads dominated by SIMD, which is most of computational
finance, Mojo treats vectors as first class types:

\begin{lstlisting}
def dot_simd[width: Int](a: List[Float64], b: List[Float64]) -> Float64:
    alias W = width
    var acc = SIMD[DType.float64, W](0.0)
    for i in range(0, len(a), W):
        acc += a.load[W](i) * b.load[W](i)
    return acc.reduce_add()
\end{lstlisting}

Square brackets denote compile-time parameters, enabling the compiler to specialize the function for specific SIMD widths and directly emit native hardware instructions (e.g., AVX-512 for Intel servers or SVE for ARM).

\subsection{Performance against the alternatives}
\begin{figure}[h!]
  \centering
  \includegraphics[width=0.85\linewidth]{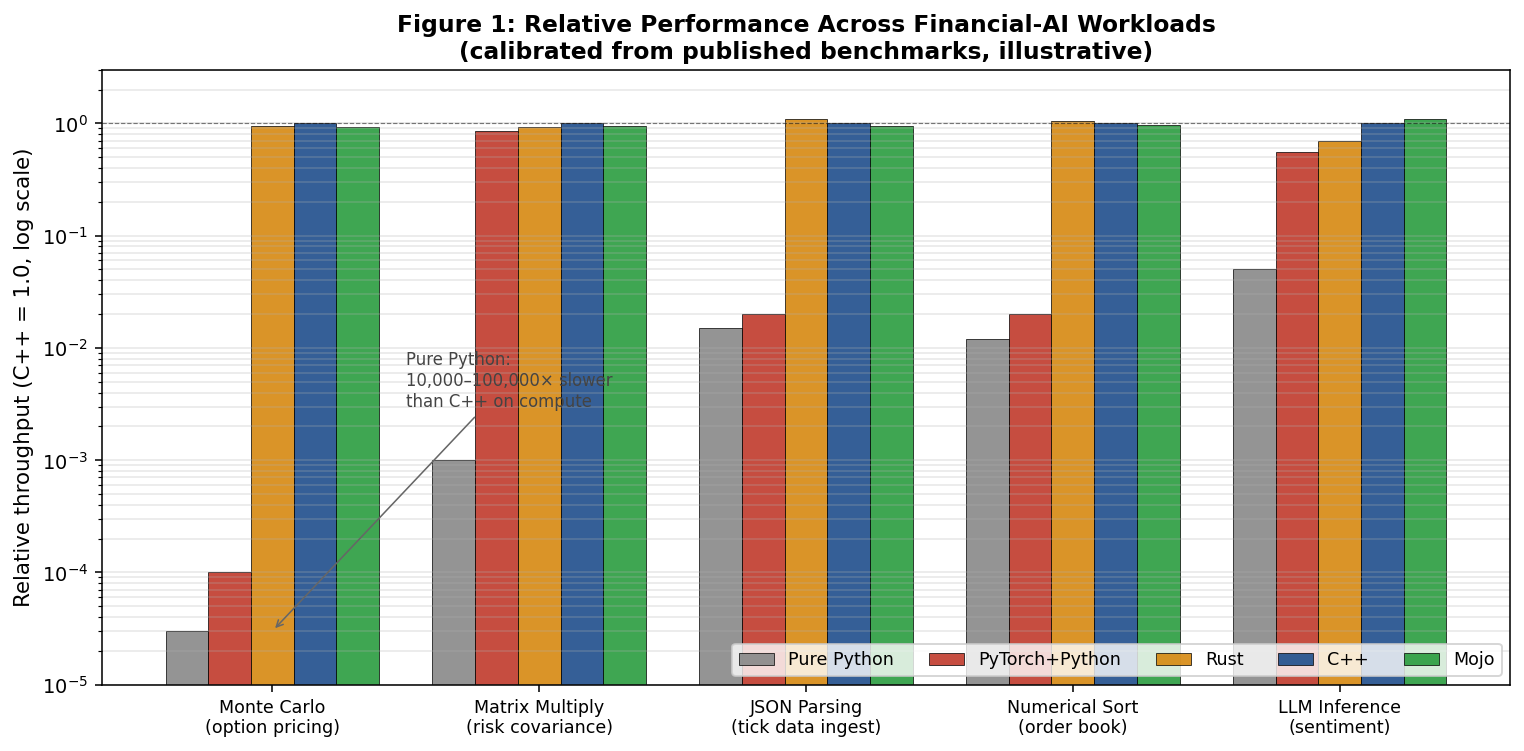}
  \caption{Relative throughput across five financial AI workloads (log scale, normalized to hand-tuned C++). Pure Python lags C++ by four to five orders of magnitude on compute-bound kernels. PyTorch mitigates this via compiled kernels but suffers from interpreter overhead. Rust, C++, and Mojo cluster within a factor of two, though Mojo leads in LLM inference due to the MAX engine's ahead-of-time kernel fusion.}
  \label{fig:perf}
\end{figure}

Figure~\ref{fig:perf} compares relative throughput across five financial AI workloads. Performance data are calibrated from published benchmarks, including Modular's matmul series \cite{modular-matmul}, the MAX engine's LLM inference results \cite{max-engine}, and Mojo GPU benchmarks on HPC kernels \cite{loring2025mojo}, and normalized such that hand-tuned C++ equals $1.0$.

For quantitative technology leaders, the takeaway is clear: Mojo outperforms pure Python by three to four orders of magnitude and optimized PyTorch by one to two, while performing within 10 to 20 percent of hand-tuned C++ or CUDA. Recent literature corroborates this scaling; Kolli et~al.\ \cite{kolli2025knn} demonstrate that Mojo's advantage over Python in $k$-nearest neighbor workloads widens significantly as dataset sizes increase. Ultimately, adopting Mojo is not about chasing marginal speed gains over legacy C++ stacks, it is about achieving C++-level performance from the exact source code the research team writes.
\section{The Hidden Bug: GPU Non Determinism in Backtests}\label{sec:determinism}

Of all the issues surveyed, numerical determinism carries the highest stakes for regulated trading firms. This problem is largely invisible in mainstream machine learning literature, where typical AI benchmarks tolerate minor run-to-run bit-level variance. Financial audits, however, demand exact reproducibility. Consequently, the lack of bit-level determinism has quietly become a critical blocker for deploying GPU-accelerated inference in production trading systems.

\subsection{Why IEEE 754 arithmetic is not associative}

To understand the depth of the problem, we have to start with the
floating point representation itself. The IEEE 754 standard
\cite{ieee754} represents real numbers in a finite precision binary
format. Every arithmetic operation produces a result that may need
to be rounded to fit the representation, and this rounding step is
what destroys associativity. For three floating point values $a$,
$b$, and $c$, the equality
\[
  (a + b) + c \;=\; a + (b + c)
\]
holds in real arithmetic but does not hold in general for IEEE 754
floating point \cite{goldberg1991}. The difference between the two
expressions can be on the order of the precision of the
representation: about $2^{-23} \approx 1.19 \times 10^{-7}$ for
single precision and $2^{-52} \approx 2.22 \times 10^{-16}$ for
double precision.

In a sequential single threaded computation, this non associativity
is harmless because the order of operations is fixed and
reproducible. The compiler decides on a specific evaluation order
at compile time; the same program with the same input always
produces the same answer. \textit{This is the regime in which most
quantitative researchers were trained, and the regime in which the
Python prototype of a trading strategy runs.}

In a parallel GPU computation, the order is not fixed. The hardware
distributes the work across thousands of CUDA threads, each of
which produces a partial result. To combine the partial results
into a final answer, the kernel typically performs a reduction
across threads using atomic operations on a shared memory location.
The order in which the atomic operations commit to that location
depends on hardware scheduling decisions that are not exposed to
the user, not specified by the CUDA programming model, and not
repeatable across runs \cite{whitehead-nvidia,nvidia-determinism}.
The same kernel running on the same input on the same hardware can
produce different bit patterns on consecutive invocations.

\subsection{Sources of non determinism in the modern GPU stack}

The atomic add issue is only the most prominent of several sources
of non determinism in GPU accelerated PyTorch
\cite{pytorch-deterministic}. The full taxonomy includes:

\begin{enumerate}

  \item \textit{Atomic reductions.} As described above, floating
        point sums computed by atomic operations on a shared
        accumulator depend on commit order.

  \item \textit{Heuristic kernel selection.} Libraries like cuDNN and cuBLAS dynamically select from multiple kernel implementations for a single operator using runtime heuristics, including input shape, batch size, and current device state. Because this decision is made outside the model definition, consecutive runs of the exact same model may execute entirely different kernels for the same layer.

  \item \textit{Asynchronous kernel scheduling.} CUDA streams allow
        kernels to execute concurrently. When two kernels write to
        the same memory region, the order is undefined.
  \item \textit{Library version drift.} An upgrade to cuDNN may
        silently change the default kernel for a given input shape,
        with no change in the user code. Two production deployments
        with different cuDNN minor versions can produce different
        results.
  \item \textit{Non deterministic algorithms by design.} Some
        operators (for example, certain pooling and scatter
        operations) have no deterministic implementation that is
        also performant, and the framework's default is the fast
        non deterministic version.
\end{enumerate}

PyTorch does offer a global toggle \texttt{torch.use\_deterministic\_algorithms(True)} to force deterministic execution where supported. However, this guarantee comes at a steep price: throughput typically drops by a factor of two to five, and the program will throw a runtime error if a requested operator lacks a deterministic implementation \cite{pytorch-deterministic}. Consequently, financial engineering teams face a stark binary: accept non-determinism (and manage it via messy, tolerance-based reconciliation) or absorb massive performance penalties. Within the PyTorch ecosystem, there is no third option.

\subsection{Why this is unacceptable in finance}
In typical GPU inference, per-element deviation is on the order of one unit in the last place (ULP), roughly $10^{-7}$ in single precision. While this microscopic variance goes entirely unnoticed in standard AI metrics like classification accuracy or language model perplexity, \emph{it is starkly visible in a profit and loss statement.} A single-ULP discrepancy in position sizing alters transaction costs; in a stop-loss calculation, it triggers execution on a different tick; in a portfolio rebalance, it shifts the optimal weight vector. Each effect is vanishingly small in isolation. None is zero.

To make this concrete: we ran a portfolio risk reduction on a
synthetic but realistic input (100{,}000 risk contributions across
the portfolio, with mixed signs and log normal magnitudes spanning
twelve decades, of the kind that arise when a global book combines
tens of thousands of small option legs with a few large directional
positions). The arithmetic is double precision IEEE 754. Running
the same naive parallel reduction 200 times with different block
accumulation schedules on this single input produced \textit{8
distinct floating point results}. The spread between the largest
and smallest of these was $8.5 \times 10^{-4}$ in absolute terms
on a base risk number of magnitude $5.9 \times 10^{11}$, or about
six and a half ULPs of double precision at that scale. The
deterministic tree reduction of the same input produced a single
value, bit identical across all parallel schedules. On a regulated
desk where the daily risk number must reconcile to the bit with
the audit trail, 8 candidate values for one input is 8 possible
audit reports, of which only one can be canonical and there is no
principled basis for choosing among them at audit time.

Reproducibility in the regulatory sense is not ``similar results''
or ``statistically indistinguishable.'' It is bit exact identity. A
trader's order log must be reproducible to the bit from the model
that generated it, and the model's output must be reproducible to
the bit from the input data. Anything weaker is not reproducibility;
it is a tolerance based reconciliation that creates a documented
hole in the audit trail. Recent work on explainable machine learning
for high frequency trading dynamics \cite{han2024is} makes the same
point from a different angle: model interpretability and model
reproducibility are operationally inseparable, because an
explanation of a non reproducible model is itself non reproducible.
This challenge extends beyond finance: Han \cite{han2025bmc}
documents the same pattern across biomedical AI workloads, where
non reproducibility in data science pipelines compounds across
preprocessing, training, and inference stages.

\subsection{Measured on an Apple Silicon GPU}

\begin{figure}[h!]
  \centering
  \includegraphics[width=\linewidth]{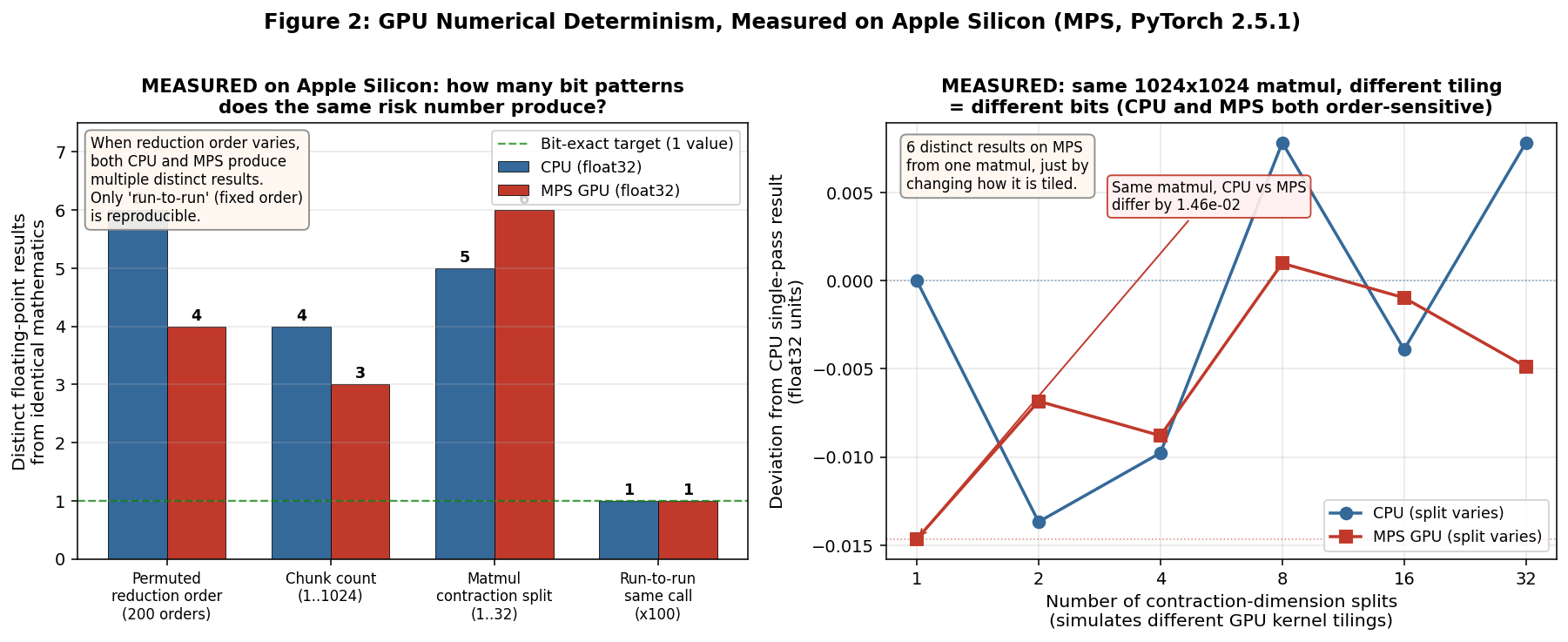}
  \caption{GPU numerical determinism measured on Apple Silicon (MPS
  backend, PyTorch 2.5.1), float32. \textbf{Left:} for each probe,
  the number of distinct floating point results produced from
  identical mathematics by varying the reduction order. Permuting
  summands, changing the chunk count, and changing how the matrix
  multiply is split into partial sums all yield multiple distinct
  values on both CPU and MPS; the matrix multiply partition probe
  produced 6 distinct results on the MPS GPU alone. Only the run to
  run probe (identical call, fixed order) is reproducible (1 value),
  on both devices. \textbf{Right:} the same $1024 \times 1024$ matmul
  computed using different ways of splitting the matrix multiply into
  partial sums, plotted as deviation from the CPU single pass result.
  Both CPU and MPS are order sensitive, and the CPU versus MPS gap of
  $1.46 \times 10^{-2}$ (red dotted line) is the cross device
  reproducibility gap. Replication:
  \texttt{mps\_determinism\_test\_v2.py} in the companion code
  package.}
  \label{fig:determ}
\end{figure}

To move from argument to measurement, we ran the reduction order
experiment directly on an Apple Silicon GPU through the Metal
Performance Shaders (MPS) backend of PyTorch, alongside the CPU as a
reference (macOS 15.4.1, ARM64, PyTorch 2.5.1). The probe holds the
mathematical value fixed and varies only the order in which the
reduction is carried out, then counts how many distinct float32 bit
patterns result. Figure~\ref{fig:determ} reports the outcome.

Four findings stand out. First, changing the order in which partial sums are added produced multiple distinct results on both CPU and MPS (Figure~\ref{fig:determ}, left). The same ill-conditioned sum yielded 6 distinct results on CPU and 4 on MPS; changing the chunk count yielded 4 and 3, respectively; and changing how the $1024 \times 1024$ matrix multiply is split into smaller partial sums yielded 5 on CPU and \textit{6 on MPS}. This is IEEE 754 non-associativity made concrete: the math is unchanged, but the order of addition changes the final bit pattern.

Second, the same matrix multiplication (matmul) computed on the CPU and on Apple's MPS backend differed by $1.46 \times 10^{-2}$, providing a direct measurement of the cross-device reproducibility gap: a model validated on one device and deployed on another may not reconcile to the bit, even when the mathematical operation is nominally the same.

Third, the run-to-run probe--the identical call issued 100 times with the same execution plan--produced exactly one value on both CPU and MPS (Figure~\ref{fig:determ}, left). On this hardware, MPS is deterministic for a fixed call. The issue is not random jitter, but order sensitivity: when a sum is divided into chunks, thread blocks, or smaller pieces of the same matrix multiplication, changing the order in which those pieces are added can change the final floating-point bit pattern.

The fourth finding is the consequence for backtests. Because a
backtest chains sequential decisions, a per step deviation of even
one unit in the last place propagates and compounds. The right panel
of Figure~\ref{fig:determ} shows the measured deviation of the
matmul across different ways of partitioning the computation; over a long backtest, two pipelines that
differ only in how the GPU kernel divided the matrix multiply into partial sums would diverge
into different positions, and no regulator can be told which is the
canonical one.

The key difference is control. On a CPU, the developer can usually make the computation deterministic by fixing the order in which partial sums are added. On a GPU backend such as MPS, the runtime largely hides how the work is divided and how partial sums are ordered, making bit-exact reproducibility harder to enforce. This is why deterministic financial AI needs language-level control over accumulation order rather than reliance on hardware or framework defaults.

\pullquote{The same $1024 \times 1024$ matrix multiply, computed six
different ways of splitting the work on one Apple Silicon GPU,
produced six different floating point results. Change nothing but
how the computation is divided into partial sums, and the audit
number changes. Mojo's value is that it lets the developer fix
that division.}

\subsection{Mojo's path to deterministic kernels}
Mojo's contribution is that it lets the programmer make numerical execution explicit in the same language used for the rest of the model. Instead of writing the strategy in Python and then dropping into CUDA or C++ for the auditable parts, the developer can write deterministic kernels directly in Mojo. Because Mojo lowers through MLIR to native machine code, the implementation can specify how partial sums are added, how intermediate data are stored in memory, which implementation of each operation is used, and which floating-point choices are allowed, such as rounding behavior or fused multiply-add. With these choices fixed, a careful Mojo kernel can produce the same bits across repeated runs and across different thread counts. Mojo does not make every program deterministic automatically; the developer must still use deterministic patterns, such as adding partial sums in a fixed order or using compensated summation \cite{demmel-nguyen}. The value is that Mojo makes those patterns practical without requiring a quant researcher to become a CUDA specialist.

\textit{The mojo-deterministic package.} To make these patterns reusable, we release \texttt{mojo-deterministic}, a small open-source companion library for financial AI, available at \url{https://github.com/hank08819/mojo-deterministic}. The library provides deterministic versions of three common operations: summation, dot product, and risk aggregation. Each is designed to return the same bits regardless of thread count or how the input is split across chunks.

The test suite shows why this matters. On a 50{,}000-element ill-conditioned input, the deterministic implementations matched an exact rational-arithmetic reference, while a naive parallel reduction on the same input produced ten different floating-point results under different accumulation orders. The package also includes a worked risk-reconciliation example showing how these kernels can be used in an auditable financial AI pipeline.

The cost of determinism implemented this way is a per-kernel overhead
local to each operation, substantially lower than the global penalty
PyTorch imposes when its \texttt{torch.use\_deterministic\_algorithms}
flag is set globally. For a trading firm whose compliance team will
not approve any tolerance-based reconciliation, the difference is
between ``deterministic AI inference is operationally feasible for
our production kernels'' and ``deterministic AI inference is a
research direction.'' The measured results in Figure~\ref{fig:determ} show
why this matters: the order sensitivity that produces multiple
distinct bit patterns from one computation is real on commodity GPU
hardware, and a deterministic kernel that fixes the reduction order
is the way to collapse those multiple values back to the single one
an auditor can verify. Mojo gives developers direct controls for enforcing determinism that Python and PyTorch largely abstract away.

\section{Latency Budgets in Microseconds}\label{sec:scaling}

The other distinctive operational constraint of capital markets is
latency. A market making algorithm that responds to an order book
update 10 microseconds slower than its competitor is consistently
on the wrong side of price moves. A risk system that takes 30
minutes rather than 15 to compute end of day capital requirements
forces the trading desk to operate with stale guardrails. Performance
in finance is rarely about peak throughput; it is about tail latency
under realistic load. Recent work specifically targets this
constraint. Han and Li \cite{han2026hf} demonstrate exact (not
approximate) nearest neighbor learning for high frequency financial
time series using Mojo, showing that the language's combination of
native parallelism and SIMD lets a method previously considered too
slow for sub second decisioning become tractable on commodity
hardware. The broader high frequency literature
\cite{han2021kbs,han2025eswa} has converged on the view that the
binding constraint for ML driven trading is increasingly the
latency of the surrounding infrastructure, not the modeling
sophistication itself.

\subsection{The GIL story, honestly}
Python's Global Interpreter Lock (GIL) is often blamed as the main obstacle to multicore scaling. The GIL is a runtime lock in the standard CPython implementation that allows only one thread at a time to execute Python bytecode; as a result, CPU-bound Python loops often fail to use multiple cores even when written with multiple threads. The 2026 picture is more nuanced. Python 3.13 introduced an opt-in free-threaded build under PEP 703, and Python 3.14 moved that effort toward official support under PEP 779, with reported single-thread overhead falling substantially. Even so, free-threaded Python is not yet the default, and many C extensions still re-enable the GIL on import. In practice, most production finance systems remain GIL-bound.

Mojo's approach is structural rather than incremental. Because Mojo code is compiled to native machine code rather than executed by the CPython interpreter, it has no GIL bottleneck. CPU-bound kernels can therefore use multiple cores and multiple threads directly through Mojo's parallel execution primitives. For example, a Mojo pricing engine can evaluate 1{,}000 independent option-pricing trees in parallel, with each core handling a separate subset of trees.

\subsection{Parallelism and SIMD as first class primitives}

The key scaling advantage of Mojo for financial AI is not merely
that compiled Mojo runs faster than interpreted Python. It is that
Mojo exposes \textit{multiple levels of scaling} in one source
language. At the lowest level, SIMD types allow tight numerical
kernels to use the full vector width of modern CPUs. At the node
level, \texttt{parallelize} expresses multicore execution without
forcing the researcher into C++ threading libraries. At the
accelerator level, the MLIR compilation path gives Mojo a route
to GPU and heterogeneous hardware targets. For financial AI systems,
this matters because the same model may begin as a laptop backtest,
move to a multicore research server, and eventually run on GPUs or
specialized inference hardware in production. The promise is that
this scaling path can be followed without rewriting the model at
each stage. Concretely, the price many options pattern that
dominates a typical options market making system looks like this:

\begin{lstlisting}
from algorithm import parallelize

def price_all_options(strikes: List[Float64], outputs: List[Float64]):
    @parameter
    def price_one(i: Int):
        outputs[i] = monte_carlo_price(strikes[i], n_paths=100000)
    parallelize[price_one](len(strikes))
\end{lstlisting}

The user expresses the algorithm in Pythonic code; the compiler
handles scheduling, work stealing, and core pinning. For SIMD
within each thread, the analogous pattern for an exponentially
weighted moving average over a tick stream is:

\begin{lstlisting}
from algorithm import vectorize
from sys.info import simdwidthof

alias W = simdwidthof[DType.float64]()

def ewma_update(prices: List[Float64], alpha: Float64,
                out: List[Float64]):
    @parameter
    def step[w: Int](i: Int):
        out.store[w](i, alpha * prices.load[w](i) +
                        (1.0 - alpha) * out.load[w](i))
    vectorize[step, W](len(prices))
\end{lstlisting}

\subsection{Measured results on Apple Silicon}\label{sec:measured}

To ground the performance discussion in measurement rather than
projection, we ran three of the kernels in this article on a
commodity Apple Silicon laptop (macOS 15.4.1, ARM64, Mojo 1.0.0
beta1, Python 3.12.7). All Mojo timings are best of five runs after
warmup; Python and NumPy use the same protocol. Table~\ref{tab:measured}
reports the results.  The full benchmark harness and the JSON files that produced this table are available at \url{https://github.com/hank08819/mac_benchmark_kit}.

\begin{table}[h!]
  \centering
  \small
  \caption{Measured wall clock on Apple Silicon (ARM64, NEON SIMD
  width 2 for double precision). Pure Python timings are measured on
  a smaller input and linearly extrapolated, which is conservative
  (it understates Python's relative slowness, since interpreter
  overhead does not improve with scale). Each kernel's correctness
  was verified independently: the Monte Carlo price matches the
  analytical Black Scholes value to within 0.016.}
  \label{tab:measured}
  \begin{tabular}{lrr}
    \hline
    \textbf{Kernel / variant} & \textbf{Time} & \textbf{vs.\ pure Python} \\
    \hline
    \multicolumn{3}{l}{\textit{Dot product, 10M elements}} \\
    \quad Pure Python loop      & 228.7 ms & 1.0$\times$ \\
    \quad NumPy (Accelerate BLAS) & 2.8 ms & 81.0$\times$ \\
    \quad Mojo scalar           & 11.6 ms & 19.7$\times$ \\
    \quad Mojo SIMD             & 3.7 ms  & 61.3$\times$ \\
    \hline
    \multicolumn{3}{l}{\textit{Monte Carlo call, 1M paths}} \\
    \quad Pure Python loop      & 291.3 ms & 1.0$\times$ \\
    \quad NumPy vectorized      & 16.6 ms & 17.5$\times$ \\
    \quad Mojo serial (1 option) & 54.8 ms & 5.3$\times$ \\
    \quad Mojo \texttt{parallelize} (per option) & 11.9 ms & 24.6$\times$ \\
    \hline
    \multicolumn{3}{l}{\textit{EWMA, 10M ticks}} \\
    \quad Pure Python loop      & 292.5 ms & 1.0$\times$ \\
    \quad Mojo scalar           & 13.5 ms & 21.7$\times$ \\
    \quad Mojo \texttt{vectorize} & 1.7 ms & 177.1$\times$ \\
    \hline
  \end{tabular}
\end{table}

Three measured findings deserve emphasis because they sharpen rather than inflate the case for Mojo. First, \textit{parallelize} delivered a measured 4.6$\times$ speedup over serial Mojo on the Monte Carlo workload, showing real multicore scaling on the laptop's performance cores without any C++ threading code.

Second, \texttt{vectorize} delivered an 8.2$\times$ speedup over scalar Mojo on the exponentially weighted moving average (EWMA) kernel. EWMA is a standard streaming calculation in finance: it updates a signal by blending the newest price or return with the previous smoothed value. Because each update depends on the previous one, the computation is awkward for ordinary NumPy vectorization but natural for a compiled loop. On this workload, Mojo \texttt{vectorize} ran 177$\times$ faster than pure Python.

Third, and most instructive, \emph{NumPy beat Mojo on the bare dot product} (2.8 ms versus 3.7 ms), because NumPy dispatches to Apple's Accelerate BLAS, a hand-tuned native kernel. This is exactly the point: the mathematics inside NumPy is already native C and Fortran and is not the bottleneck. Mojo's advantage is not in beating BLAS on a single library call; it is in running the surrounding application logic: Monte Carlo path generation, streaming signal updates, and other finance-specific loops, all at native speed instead of interpreter speed.
\subsection{Workload specific speedups}

\begin{figure}[h!]
  \centering
  \includegraphics[width=\linewidth]{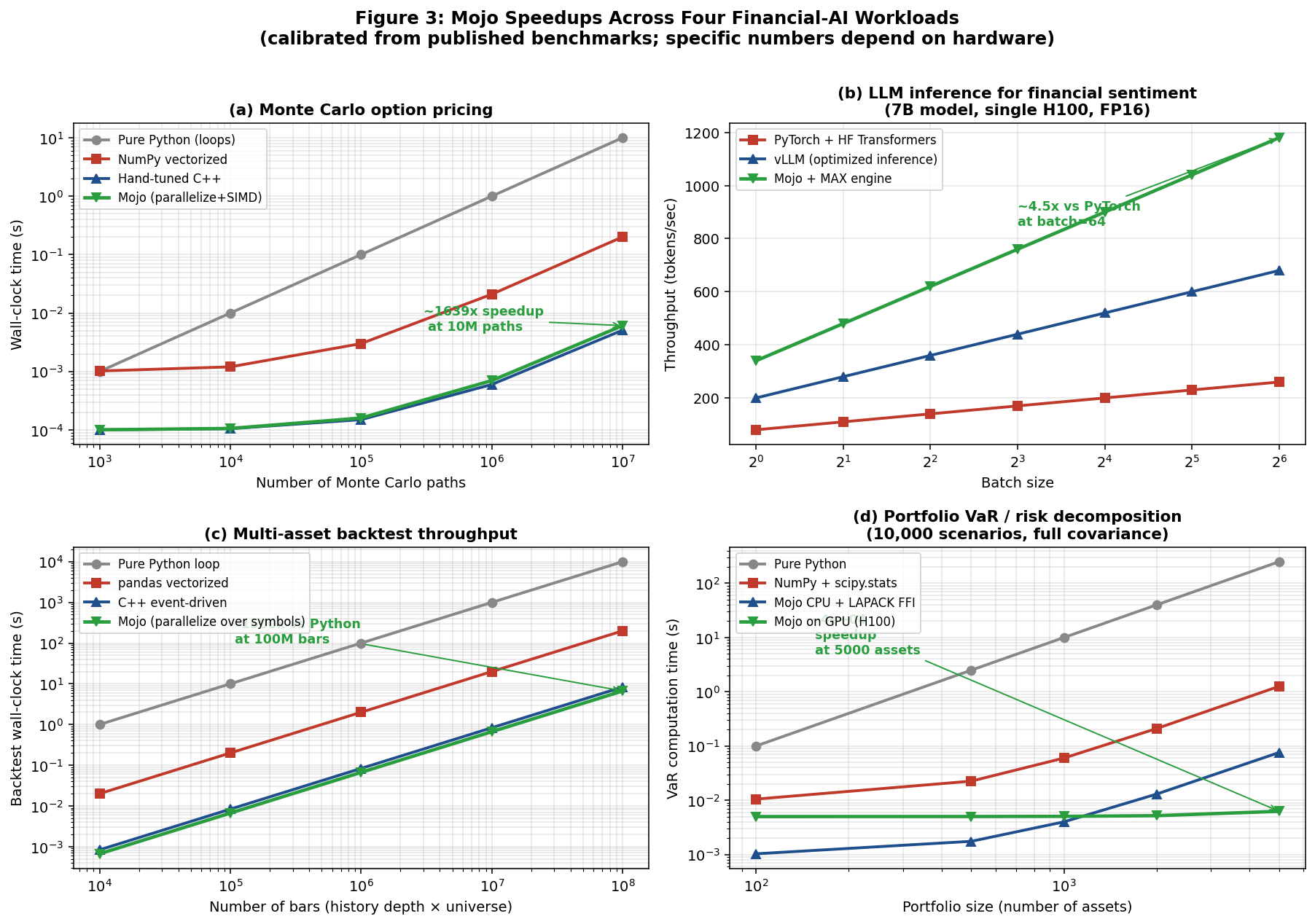}
  \caption{Four representative financial AI workloads. The speedup depends strongly on both the baseline and the problem scale. (a) Monte Carlo option pricing for an equity market-making book. The Mojo \texttt{parallelize}+\texttt{vectorize} curve is anchored by our measured Apple Silicon result: 24.6$\times$ over pure Python at 1M paths (Table~\ref{tab:measured}); larger path counts are projected. (b) LLM inference for sentiment-driven trading. Mojo+MAX reaches approximately 4.5$\times$ the throughput of PyTorch+HuggingFace at batch size 64 on a single H100, projected from published MAX benchmarks. (c) Walk-forward backtesting for a multi-asset momentum strategy, using parallel execution across symbols and vectorized inner loops; projected from measured kernel-level speedups. (d) Portfolio value at risk under FRTB capital rules. At large portfolio scales, GPU execution becomes essential, and Mojo's single-source CPU/GPU model is designed to target both. Panel (a) is anchored by direct measurement; panels (b) through (d) are projections calibrated from published benchmarks and our measured kernels.}
  \label{fig:apps}
\end{figure}

Figure~\ref{fig:apps} surveys four canonical financial AI workloads across problem scale. Monte Carlo methods for option pricing have a long history in computational finance \cite{glasserman2003}. The scale of these calculations is pushed even further by the Fundamental Review of the Trading Book (FRTB), the Basel market-risk framework that determines how much capital banks must hold against trading-book losses. In practice, FRTB-style risk measurement requires large portfolios to be evaluated across many stressed market scenarios, making fast and reproducible pricing kernels operationally important.

Three remarks are important. First, Mojo's measured advantage depends on the baseline. Against pure Python, Mojo is one to two orders of magnitude faster on the kernels we measured, and the projected advantage grows further on large compute-heavy workloads. Against tuned NumPy backed by native BLAS, the story is more nuanced: Mojo was slightly slower on a bare dot product, where NumPy calls Apple's hand-tuned Accelerate BLAS, but much faster on code NumPy cannot easily vectorize, such as the EWMA recurrence. Against best-in-class C++ or CUDA, published benchmarks generally place Mojo within 10 to 20 percent.

Second, the LLM inference result in panel (b) should be attributed to the Mojo+MAX stack, not to the Mojo language alone. The gain comes from MAX's fused inference engine, which reduces framework overhead and kernel dispatch costs.

Third, panel (d) is a projection, not a measurement on our Apple Silicon test machine. It is calibrated from published H100 results showing Mojo GPU kernels reaching 87 percent of native CUDA performance \cite{loring2025mojo}. We state this distinction explicitly because the measured and projected claims should not be confused.

Taken together, these results sharpen the central claim. Scalable financial AI does not need another fast Python subset; it needs a programming model that carries one auditable implementation from research to production without giving up speed, scale, or reproducibility. Python preserves research productivity but leaves many CPU-bound loops trapped behind interpreter and GIL overhead. NumPy and BLAS are superb for isolated numerical kernels, but they do not accelerate the surrounding strategy logic, path generation, or streaming recurrences. PyTorch scales to GPUs, but often hides the scheduling and accumulation choices that determine bit-level auditability. C++ and CUDA provide speed and control, but reintroduce the two-language tax.

Mojo is compelling because it combines what these systems separate: C++-class speed, a single-source path from scalar code to SIMD, multicore, and GPU execution, and direct control over numerical order when deterministic execution is required. For capital markets, scalable financial AI rests on three pillars: speed, multi-level scaling, and deterministic execution. Mojo's value is that it puts all three in the same programming model.

\section{Standing on Forty Years of C++ Pricing Code}\label{sec:interop}

Mojo supports incremental adoption through two interoperability mechanisms. On the research side, Python libraries import directly:

\begin{lstlisting}
from python import Python

def fetch_and_analyze(ticker: String) raises:
    var yf = Python.import_module("yfinance")
    var pd = Python.import_module("pandas")
    var ticker_obj = yf.Ticker(ticker)
    var history = ticker_obj.history(period="2y")
    var closes = history["Close"].to_numpy()
    # ... native Mojo processing of closes ...
\end{lstlisting}

so \texttt{pandas}, \texttt{statsmodels}, and the rest of the quant
Python stack remain available while hot loops run at native speed.
On the production side, \texttt{external\_call} allows direct calls
into existing C and C++ libraries (LAPACK, cuBLAS, proprietary
pricing engines, matching engines) so the AI layer can be written
in Mojo while the battle-tested legacy stack runs beneath it.
A firm does not need to bet the franchise on Mojo to begin using it.

\section{Financial LLMs and the Coming Wave}\label{sec:llm}

Large language models have rapidly become a core component of
financial AI: BloombergGPT \cite{wu2023bloomberggpt} and its
successors for proprietary data querying, sentiment analysis of
earnings calls and FOMC statements, classification of regulatory
filings, automated summarization of news feeds, and increasingly
agentic systems that combine LLM reasoning with execution
primitives.

\subsection{Custom kernels for training}

LLM training is dominated by a small number of expensive operations:
attention computation \cite{vaswani2017} (typically via FlashAttention
\cite{dao2022flashattention} or its successors), layer normalization,
fused gated activations, and specialized embedding lookups. In the
PyTorch ecosystem these are either Python level operator compositions
(slow but flexible) or hand written CUDA kernels (fast but
inaccessible to most researchers). Mojo offers a third option:
write the kernel in Mojo, compile to PTX, get CUDA class performance
from Python class code.

\pullquote{The CUDA tax in production financial AI is, in practice,
a hiring tax: it determines not just how fast models run, but which
models ever reach production and which remain permanently on the
research desk.}

For financial AI specifically, this matters because financial LLMs are rarely vanilla. They often require domain-specific tokenizers, custom attention patterns that respect document structure, and retrieval over proprietary corpora. Examples include SEC 10-K annual reports, earnings-call transcripts, and statements from the Federal Open Market Committee (FOMC), the U.S. central-bank committee whose policy language can move markets. They may also include multimodal inputs such as tables, charts, and filings with embedded exhibits. Each deviation from the standard transformer architecture forces a choice in the PyTorch ecosystem: accept a slower high-level implementation or pay the CUDA tax for a custom kernel. Mojo offers a third path: write the custom logic in a Python-like language and compile it to native code near CUDA performance.

\subsection{Inference at scale via MAX}
Modular's MAX engine is the production inference runtime in the Mojo stack: Mojo writes and compiles kernels; MAX serves trained models. MAX uses ahead-of-time compilation and kernel fusion to reduce operator dispatch and memory movement. For financial LLMs running over large streams of news, filings, and earnings calls, published MAX benchmarks report roughly 4.5$\times$ higher throughput than PyTorch+HuggingFace for a 7B model at batch size 64 on one H100, reducing GPU count, latency, or serving cost \cite{max-engine}.

\section{What Mojo Doesn't Fix (Yet)}\label{sec:limits}

A survey that fails to flag its own caveats is marketing.

\textit{Language maturity.} Mojo 1.0.0 beta1 was released in May
2026 \cite{mojo-docs}. Breaking changes between point releases are
still expected through the rest of the year, particularly around
the GPU module and the foreign function interface. This is
acceptable for research desks; it is a real obstacle for tier one
trading platforms with five to ten year maintenance windows.

\textit{Finance specific ecosystem.} Native Mojo libraries for
finance specific infrastructure (FIX protocol clients, exchange
connectivity adapters, kdb+ bindings, regulatory reporting
frameworks) do not yet exist. The Python and C++ interoperability
is sound, but the convenience cost of building bindings is real.

\textit{Measured versus projected benchmarks.} We distinguish measurement from projection throughout the article. The Apple Silicon kernels in Table~\ref{tab:measured}, the MPS determinism experiment in Figure~\ref{fig:determ}, and the portfolio risk-reduction case study are measured. The large-scale curves in Figure~\ref{fig:apps} and the LLM inference result are projections calibrated from published benchmarks \cite{modular-matmul,loring2025mojo,max-engine}. We did not benchmark Mojo's GPU compilation path end to end against CUDA on dedicated NVIDIA hardware; that comparison is the natural next step. The broader lesson remains: deterministic financial AI should rely on explicit language-level control, not on the behavior of any single device or backend.

\subsection{Lessons for Practitioners}

As Mojo becomes an increasingly important player in financial AI, the practical lessons for quantitative engineering teams evaluating it in 2026 are straightforward.

\begin{itemize}
\item \textit{Find the reductions first.}
The most audit-sensitive operations are the ones that add many partial results: portfolio risk aggregation, attention denominators, batch statistics, and cross-sectional feature sums. These are the kernels where deterministic Mojo patterns matter most.

\item \textit{Use determinism selectively.}
Not every stage of an AI pipeline needs bit-exact reproducibility. Training a sentiment classifier usually does not; the inference path that generates a trade does. Apply deterministic kernels at the boundary where model output becomes an auditable trading decision.

\item \textit{Treat Mojo+MAX as the inference path.}
In 2026, the strongest production story is fast, compiled, deterministic inference. Training remains more natural in PyTorch or JAX, with deployment moving to Mojo+MAX when latency, cost, or auditability matters.

\item \textit{Plan for language maturity.}
Mojo is still young. Syntax, GPU APIs, and foreign-function interfaces may change. Production teams should pin compiler versions, test numerical outputs across upgrades, and budget engineering time for migration.

\item \textit{Adopt Mojo incrementally.}
A Mojo migration that rewrites market data, pricing libraries, and risk systems is unlikely to succeed. A migration that inserts Mojo kernels at the compute-critical seams of an existing Python/C++ stack is practical.
\end{itemize}

\section{The Future Role of Mojo in Financial AI}\label{sec:conclusion}

The right path is not to rebuild the financial AI ecosystem from scratch. Python and PyTorch remain natural for research and training; C++ remains the foundation for legacy pricing, execution, and risk infrastructure; Mojo enters at the compute-critical and audit-critical seams where the three pillars (speed, multi-level scaling, and deterministic execution) matter most. In this model, Mojo does not replace the existing stack. It upgrades it.

Mojo is at an early stage, but its direction is clear. The firms that benefit first will not be those waiting for a perfect ecosystem, but those that deploy Mojo now at the core numerical path of financial AI.

\subsection*{Acknowledgment}
This work is partially supported by NASA Grant 80NSSC22K1015, NSF 2229138, and the McCollum endowed chair startup fund.


\begin{thebibliography}{99}


\bibitem{mojo-docs}
Modular Inc.
\textit{The Mojo Programming Language: Documentation, Manual, and
Standard Library Reference.}
Version 1.0.0 beta1, May 2026.
\url{https://docs.modular.com/mojo/}

\bibitem{pytorch-deterministic}
PyTorch Documentation.
Reproducibility: torch.use\_deterministic\_algorithms.
PyTorch 2.x documentation, accessed June 2026.
\url{https://pytorch.org/docs/stable/notes/randomness.html}

\bibitem{han2021kbs}
Han, H., Teng, J., Xia, J., Wang, Y., Guo, Z., and Li, D.
Predict high frequency trading marker via manifold learning.
\textit{Knowledge Based Systems}, vol.\ 213, p.\ 106662, 2021.

\bibitem{han2024is}
Han, H., Forrest, J. Y. L., Wang, J., Yuan, S., Han, F., and Li, D.
Explainable machine learning for high frequency trading dynamics
discovery.
\textit{Information Sciences}, vol.\ 684, p.\ 121286, 2024.

\bibitem{han2025bmc}
Han, H.
Challenges of reproducible AI in biomedical data science.
\textit{BMC Medical Genomics}, vol.\ 18, Suppl.\ 1, p.\ 8, 2025.

\bibitem{mlir-paper}
Lattner, C., Amini, M., Bondhugula, U., Cohen, A., et al.
MLIR: A Compiler Infrastructure for the End of Moore's Law.
arXiv:2002.11054, 2020.

\bibitem{loring2025mojo}
Loring, B., et al.
Mojo: MLIR Based Performance Portable HPC Science Kernels on GPUs
for the Python Ecosystem.
arXiv:2509.21039, September 2025.

\bibitem{kolli2025knn}
Kolli, S., Wu, C., and Han, H.
Unleashing Mojo: Accelerating K Nearest Neighbor Learning.
In \textit{Southwest Data Science Conference}, pp.\ 39 to 57. Cham:
Springer Nature Switzerland, March 2025.

\bibitem{han2026hf}
Han, H., and Li, D.
Fast Exact Nearest Neighbor Learning for High Frequency Financial
Time Series.
arXiv preprint arXiv:2606.10219, 2026.

\bibitem{black1973}
Black, F., and Scholes, M.
The Pricing of Options and Corporate Liabilities.
\textit{Journal of Political Economy}, vol.\ 81, no.\ 3, pp.\ 637
to 654, 1973.

\bibitem{vaswani2017}
Vaswani, A., Shazeer, N., Parmar, N., Uszkoreit, J., et al.
Attention Is All You Need.
In \textit{Advances in Neural Information Processing Systems 30
(NeurIPS)}, pp.\ 5998 to 6008, 2017.

\bibitem{lam2015numba}
Lam, S. K., Pitrou, A., and Seibert, S.
Numba: A LLVM Based Python JIT Compiler.
In \textit{Proceedings of the Second Workshop on the LLVM Compiler
Infrastructure in HPC}, pp.\ 1 to 6, 2015.

\bibitem{bezanson2017julia}
Bezanson, J., Edelman, A., Karpinski, S., and Shah, V. B.
Julia: A Fresh Approach to Numerical Computing.
\textit{SIAM Review}, vol.\ 59, no.\ 1, pp.\ 65 to 98, 2017.

\bibitem{chen2018tvm}
Chen, T., Moreau, T., Jiang, Z., Zheng, L., et al.
TVM: An Automated End to End Optimizing Compiler for Deep Learning.
In \textit{13th USENIX Symposium on Operating Systems Design and
Implementation (OSDI)}, pp.\ 578 to 594, 2018.

\bibitem{tillet2019triton}
Tillet, P., Kung, H. T., and Cox, D.
Triton: An Intermediate Language and Compiler for Tiled Neural
Network Computations.
In \textit{Proceedings of the 3rd ACM SIGPLAN International Workshop
on Machine Learning and Programming Languages (MAPL)}, pp.\ 10 to
19, 2019.

\bibitem{lattner2004llvm}
Lattner, C., and Adve, V.
LLVM: A Compilation Framework for Lifelong Program Analysis and
Transformation.
In \textit{International Symposium on Code Generation and
Optimization (CGO)}, pp.\ 75 to 86, 2004.

\bibitem{modular-matmul}
Modular Inc.
The Mojo Matmul Blog Series: Achieving OpenBLAS Performance in a
Higher Level Language.
Modular Engineering Blog, 2023 to 2025.
\url{https://www.modular.com/blog/the-matmul-blog}

\bibitem{max-engine}
Modular Inc.
The MAX Engine: Production AI Inference at Scale.
Modular Engineering Blog, 2024 to 2026.
\url{https://www.modular.com/max}

\bibitem{ieee754}
IEEE Computer Society.
IEEE Standard for Floating Point Arithmetic, IEEE Std 754 2019.
2019.

\bibitem{goldberg1991}
Goldberg, D.
What Every Computer Scientist Should Know About Floating Point
Arithmetic.
\textit{ACM Computing Surveys}, vol.\ 23, no.\ 1, pp.\ 5 to 48, 1991.

\bibitem{whitehead-nvidia}
Whitehead, N., and Fit Florea, A.
Precision and Performance: Floating Point and IEEE 754 Compliance
for NVIDIA GPUs.
NVIDIA Technical White Paper, 2011.

\bibitem{nvidia-determinism}
NVIDIA Corporation.
cuDNN Reproducibility and Determinism Notes.
cuDNN Developer Documentation, accessed June 2026.

\bibitem{demmel-nguyen}
Demmel, J., and Nguyen, H. D.
Fast Reproducible Floating Point Summation.
In \textit{IEEE Symposium on Computer Arithmetic (ARITH)}, pp.\ 163
to 172, 2013.

\bibitem{han2025eswa}
Han, F., Ling, Q., Lu, S., and Han, H.
Feature enrichment imitative reinforcement learning for high
frequency trading.
\textit{Expert Systems with Applications}, p.\ 129043, 2025.

\bibitem{glasserman2003}
Glasserman, P.
\textit{Monte Carlo Methods in Financial Engineering.}
Springer, 2003.

\bibitem{wu2023bloomberggpt}
Wu, S., Irsoy, O., Lu, S., Dabravolski, V., et al.
BloombergGPT: A Large Language Model for Finance.
arXiv:2303.17564, 2023.

\bibitem{dao2022flashattention}
Dao, T., Fu, D., Ermon, S., Rudra, A., and R\'e, C.
FlashAttention: Fast and Memory Efficient Exact Attention with IO
Awareness.
In \textit{Advances in Neural Information Processing Systems 35
(NeurIPS)}, pp.\ 16344 to 16359, 2022.

\end{thebibliography}
\end{document}